\documentclass[sigconf, nonacm, authorversion]{acmart}

\AtBeginDocument{%
  \providecommand\BibTeX{{%
    \normalfont B\kern-0.5em{\scshape i\kern-0.25em b}\kern-0.8em\TeX}}}

\acmConference[Pre-print 2024]{}{}{}

\usepackage{csquotes}
\usepackage{subfig}
\usepackage{todonotes}

\begin{document}

\title{Generative AI in Education: A Study of Educators' Awareness, Sentiments, and Influencing Factors}


\author{Aashish Ghimire}
\email{aashish.ghimire@usu.edu}
\orcid{0000-0002-6656-543X}

\affiliation{%
  \institution{Utah State University}
  \streetaddress{Old Main Hill}
  \city{Logan}
  \state{Utah}
  \country{USA}
  \postcode{84322}
}

\author{James Prather}
\email{james.prather@acu.edu}
\orcid{0000-0003-2807-6042}
\affiliation{%
  \institution{Abilene Christian University}
  \city{Abilene}
  \state{Texas}
  \country{USA}}

\author{John Edwards}
\email{john.edwards@usu.edu}
\orcid{0000-0002-0882-312X}
\affiliation{%
  \institution{Utah State University}
  \streetaddress{Old Main Hill}
  \city{Logan}
  \state{Utah}
  \country{USA}
  \postcode{84322}
}





\begin{abstract}
The rapid advancement of artificial intelligence (AI) and the expanding integration of large language models (LLMs) have ignited a debate about their application in education. This study delves into university instructors' experiences and attitudes toward AI language models, filling a gap in the literature by analyzing educators' perspectives on AI's role in the classroom and its potential impacts on teaching and learning. The objective of this research is to investigate the level of awareness, overall sentiment towards adoption, and the factors influencing these attitudes for LLMs and generative AI-based tools in higher education. Data was collected through a survey using a Likert scale, which was complemented by follow-up interviews to gain a more nuanced understanding of the instructors' viewpoints. The collected data was processed using  statistical and thematic analysis techniques. Our findings reveal that educators are increasingly aware of and generally positive towards these tools. We find no correlation between teaching style and attitude toward generative AI. Finally, while CS educators show far more confidence in their technical understanding of generative AI tools and more positivity towards them than educators in other fields, they show no more confidence in their ability to detect AI-generated work.

\end{abstract}

\ccsdesc[500]{Social and professional topics~Computing education}
\ccsdesc[300]{Applied computing~Computer-assisted instruction}
\ccsdesc[300]{Computing methodologies~Machine learning approaches}

\keywords{LLM, Chatbot, ChatGPT, AI in Education, Teachers' attitude, AI in Classroom, Generative AI in Education, Survey}



\maketitle


\section{Introduction}
The rapid advancement of generative artificial intelligence (AI) and the increasing integration of large language models (LLMs) in various domains have sparked a debate surrounding their implementation within the educational sector \cite{becker2023programming, lau2023ban, ko2023more}. This study aims to investigate instructors' experiences and attitudes toward harnessing AI language models in education, focusing on understanding the underlying factors that shapes these opinions. The study addresses a gap in the existing literature by comprehensively analyzing educators' perspectives on integrating AI technologies in the classroom and their implications for teaching and learning.

To achieve the research objectives, the study explores the following research questions:

\begin{itemize}

    \item [\textbf{RQ1}] How aware are educators of Generative AI-based tools across various departments?
   
    \item[\textbf{RQ2}] What are educators’ perceptions and sentiments about these AI tools?
    \item[\textbf{RQ3}] What factors contribute to variations in teachers’ attitudes toward generative AI based tools?
    \item[\textbf{RQ4}] How do the attitudes and perceptions of CS educators differ from those of educators in different departments?
    \item[\textbf{RQ5}] What are the biggest opportunities and concerns identified by the educators?
\end{itemize}

The study employed a mixed-methods research design, incorporating both quantitative and qualitative data collection and analysis techniques. A survey was conducted to collect data on instructors' experiences and attitudes using a Likert scale, which was supplemented by free-form text entries and optional interviews to gain a more nuanced understanding of the factors shaping their perspectives.  The data were analyzed using statistical and thematic analysis techniques to generate insights into the research questions.

By understanding educators' experiences and attitudes concerning the harnessing of AI language models in education, this study aims to contribute to the ongoing discourse on the role of AI technologies in shaping the future of education. The findings can inform policymakers, educators, and researchers about the potential benefits and challenges of integrating AI language models into the classroom and guide the development of strategies and practices that enhance teaching and learning outcomes.

 
         








\section{Related Work}

\subsection{Generative AI tools in Computer Science Education}
Recent advances in generative AI and natural language processing have enabled the development of large language models (LLMs) that show impressive capabilities in generating and reasoning about code \cite{denny2024computing}. Major LLM-based products like Generative Pre-trained Transformer (GPT-4), CodeX, GitHub Copilot, Bard and ChatGPT have significant implications for computing education research and practice\cite{prather2024robots}.

A growing body of work has begun empirically evaluating how these LLMs perform on tasks and assessments commonly used in programming courses \cite{finnieansley2022robots,finnie-ansley2023my}. For instance, Chen et al. found that GPT-3, after generating 100 samples and selecting the sample that passed the unit tests, scored around 78\% on CS1 exam questions, outperforming most students ~\cite{chen2021evaluating}. In more advanced CS2 assessments, Codex performed comparably to the students in top quartile \cite{savelka2023large}. GitHub Copilot was also shown to generate passing solutions for typical introductory programming assignments \cite{puryear2022github}. These studies clearly demonstrate the need to reconsider curriculum design and assessment in light of LLM capabilities.

Researchers have proposed adaptations such as focusing less on basic coding skills and more on higher-level thinking and analysis when LLMs can automate generation \cite{chen2021evaluating}. New forms of assessment may be required to prevent plagiarism and ensure students have true mastery \cite{finnie-ansley2023my,denny2023conversing, ross2023case}. There are also calls to explicitly teach the productive use of LLMs as aids rather than relying excessively on them \cite{ross2023case,ernst2022ai,denny2023chatoverflow}.

Beyond assessment, researchers have identified opportunities for using LLMs in pedagogy. They can automatically generate solutions, explanations, and examples to scaffold learning and reduce instructor effort \cite{macneil2022generating, sarsa2022automatic, becker2023programming, denny2022robosourcing, macneil2023experiences}. LLMs may enable novel active learning approaches through personalized help, peer code reviews, and interactive coding activities integrated with LLMs \cite{becker2023programming,leinonen2023using,prather2024its, reeves2023evaluating}. New programming problem types that utilize LLMs, such as Prompt Problems, are also beginning to emerge \cite{ denny2024prompt}. However, risks include the propagation of incorrect solutions or explanations if not vetted \cite{becker2023programming, chen2021evaluating}.

The literature also highlights threats posed by LLMs regarding over-reliance impeding learning \cite{vaithilingam2022expectation} and circumventing assessments \cite{chen2021evaluating, denny2023conversing}. Challenges around plagiarism detection \cite{puryear2022github, savelka2023large}, bias \cite{liu2023summary}, and the greater socio-economical consequences \cite{luitse2021great} must also be addressed. Further research is critically needed to develop evidence-based practices for effectively leveraging LLMs in computing courses while mitigating their potential harms.

\subsection{Teachers' attitudes towards AI tools in education}

The attitudes and perceptions of instructors and educators are paramount in the adoption, rejection, success, or failure of these tools. Bii et al. investigated the attitude of teachers towards the use of chatbots in routine teaching by surveying teachers in Kenya, and the results showed that teachers have a positive attitude towards the use of chatbots \cite{bii2018teacher}. The study found that teachers have some reservations about using chatbots, such as concerns about the accuracy of the information provided by chatbots and the potential for chatbots to replace teachers. However, overall, the study found that teachers are open to using chatbots in their teaching. Guillén-Gámez and Mayorga-Fernández investigated the factors that predict teachers' attitudes towards information and communication technologies (ICT) in higher education for teaching and research \cite{Guillén-Gámez_Mayorga-Fernández_2020}. The results of the study showed that the professors' attitudes towards ICT were positively predicted by their age, gender, and participation in ICT-related projects. The professors' attitudes were also positively predicted by their teaching experience and their perception of the usefulness of ICT for teaching and research. Nazaretsky et al. investigated the factors that influence teachers’ attitudes towards AI-based educational technology \cite{Nazaretsky_Cukurova_Ariely_Alexandron_2021}.The study found that teachers’ attitudes were influenced by two human factors: confirmation bias and trust. Teachers who were more likely to engage in confirmation bias were more likely to ignore information about AI-based educational technology that contradicted their existing beliefs, thus becoming less likely to have positive attitudes towards AI-based educational technology.
 
Akgun and Greenhow provided an in-depth exploration of the ethical challenges inherent in the deployment of artificial intelligence (AI) within K-12 educational settings \cite{Akgun_Greenhow_2022}. The authors highlight the importance of transparency, accountability, sustainability, privacy, security, inclusiveness, and human-centered design in the development and use of AI in education. Celik et al. explores the roles of teachers in AI research, the advantages of AI for teachers, and the challenges they face in using AI \cite{Celik_Dindar_Muukkonen_Järvelä_2022}. They found that teachers have seven roles in AI research, including providing data to train AI algorithms and offering input on students' characteristics for AI-based implementation. The advantages of AI for teachers were identified in planning, implementation, and assessment, with AI providing timely monitoring of learning processes and assisting in decision-making on student performance. However, the study also highlighted challenges such as the limited technical capacity of AI, the lack of technological knowledge among teachers, and the context-dependency of AI systems. Kim and Kim investigated the perceptions of STEM teachers towards the use of an AI-enhanced scaffolding system developed to support students' scientific writing \cite{Kim_Kim_2022}. The results of the study showed that the teachers had a generally positive perception of the AI-enhanced scaffolding system. The teachers felt that the system could be used to provide personalized instruction, automate tasks, and provide feedback to students. Despite the positive expectations, the study noted that before AI can be effectively adopted in classrooms, teachers first need to learn how to use this technology and understand its benefits.

Chocarro et al. recently examined the factors that influence teachers' attitudes towards chatbots in education \cite{Chocarro_Cortiñas_Marcos-Matás_2023}. They used the dimensions of the Technology Acceptance Model (TAM), specifically perceived usefulness and perceived ease of use, to understand this acceptance. The study takes into account the conversational design of the chatbot, including its use of social language and proactiveness, as well as characteristics of the users, such as the teachers’ age and digital skills. They found that formal language used by a chatbot increased teachers' intention to use them, and teachers' age and digital skills were related to their attitudes towards chatbots. Khong et al. aimed to construct a model that predicts teachers’ extensive technology acceptance by examining the factors that influence their behavioral intention to use technology for online teaching by extending  Technology Acceptance Model (TAM) \cite{Khong_Celik_Le_Lai_Nguyen_Bui_2023}. The study suggested that cognitive attitude had a much larger impact on teachers’ behavioral intention to teach online, and perceived usefulness of online learning platforms had greater influence on teachers’ online teaching attitude than perceived ease of use, particularly on cognitive attitude.

The 2023 study by Iqbal et al. explored the attitudes of faculty members towards using ChatGPT \cite{Iqbal_Ahmed_Azhar_2023}. The study used the TAM to investigate the factors that influence faculty members' attitudes towards using ChatGPT. The study found that faculty members had a generally negative perception and attitude towards using ChatGPT. Potential risks such as cheating and plagiarism were cited as major concerns, while potential benefits such as ease in lesson planning and assessment were also noted. Finally, 
Lau and Guo present the perspectives of 20 university instructors who teach introductory programming courses on how they plan to adapt to the growing presence of AI code generation and explanation tools such as ChatGPT and GitHub Copilot \cite{lau2023ban}. They report that instructors have different opinions on whether to resist or embrace these tools in their courses and propose a set of open research questions for the computing education community.

\section{Methodology}
\label{sec:methodology}

\subsection{Survey Design}
To investigate teachers' attitudes toward AI tools and Language Learning Models (LLMs) in education, we conducted a quantitative study using a survey. The survey was designed to explore educators' perceptions of AI language models and their integration into pedagogical practices. It included questions that assessed participants' awareness of AI and LLMs, their beliefs about the potential benefits and challenges of these technologies, and their attitudes toward using them in the classroom.

The survey questions were developed based on relevant literature and the research questions listed above. The Likert scale was used for most questions, allowing participants to indicate their level of agreement or disagreement with specific statements. Additionally, the survey included open-ended questions to capture qualitative insights and gather more in-depth responses as well as some basic anonymous demographic data like age-group and tenure length. The survey was IRB approved at [anonymous].

\subsection{Data Collection}
We distributed the survey to faculty members at a mid-sized research university in the western United States via email. Each faculty member received the survey link only once to avoid duplicate responses. The email provided a brief introduction to the research study, assured confidentiality, and encouraged participation. Participants were informed about the voluntary nature of the survey and were given the option to opt-in for a follow-up interview.

\subsection{Survey Responses}
We received a total of 116 survey responses from email requests, representing a diverse sample from 8 colleges and 23 out of 39 departments at the university. The wide-ranging representation ensures a comprehensive understanding of educators' attitudes from various academic disciplines.

\subsection{Interviews}
To gain deeper insights into teachers' experiences and attitudes, we conducted semi-structured interviews with a subset of participants who opted-in for the follow-up interview. The interviews were approximately 25-30 minutes long and used open-ended questions to encourage participants to share their perspectives freely. The interview responses were recorded and later transcribed for analysis. Interviews were also conducted with IRB oversight.

\subsection{Data Analysis}
\subsubsection{Quantitative Study}
The quantitative survey data were analyzed using descriptive statistics and inferential methods. We calculated the mean, standard deviation, and frequency distributions to summarize participants' responses to Likert scale questions. We also performed hypothesis tests, confidence intervals, and regression analysis to identify potential correlations and trends in the data.

\subsubsection{Grounded Theory and Qualitative Analysis}
For the qualitative analysis, we adopted a grounded theory~\cite{lazar2017research} approach to identify themes and patterns emerging from the interview data. Three independent evaluators coded two transcribed interviews each, and inter-rater reliability was evaluated using Cohen's Kappa coefficient. The evaluation resulted in an inter-rater reliability of over 85\%, ensuring the consistency of the coding process.

\subsubsection{Integration of Data}
The coded interview data were integrated with the quantitative survey results to triangulate findings and provide a comprehensive understanding of teachers' attitudes toward AI tools and LLMs. The grounded theory approach allowed us to generate inductive insights from the interview data, which were then co-analyzed with the quantitative study's results.

\subsection{Participation}
\begin{figure}[h]
\includegraphics[width=8cm]{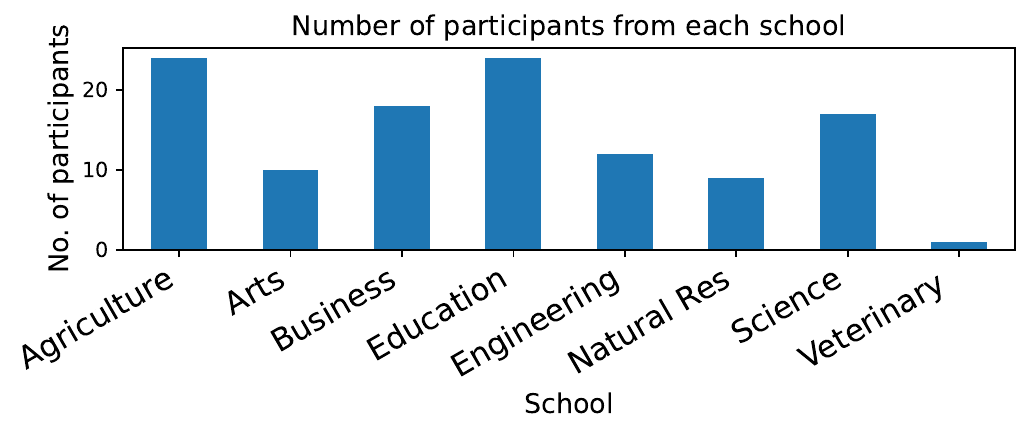}
\caption{Participants across various schools.}
\label{fig:participents}
\end{figure}

The survey received a total of 116 responses from faculty members across various school/colleges and departments at the university. The colleges with the highest number of respondents were the College of Arts and Sciences (Science), the College of Education and Human Services (Education), and the School of Business (Business). In terms of follow-up interviews, 36 faculty members opted for further discussions, with a notable interest from the College of Science (Science) and the College of Engineering (Engineering). This diverse representation of faculty members provides a broad perspective on the attitudes and opinions towards AI tools and LLM-based technologies in education. Figure \ref{fig:participents} shows the participants from each school and department. 

\section{Results and discussion}


\begin{figure}
    \centering
    \subfloat[]{
    \label{fig:familiarity_with_llms_by_school}
    \includegraphics[width=0.48\columnwidth]{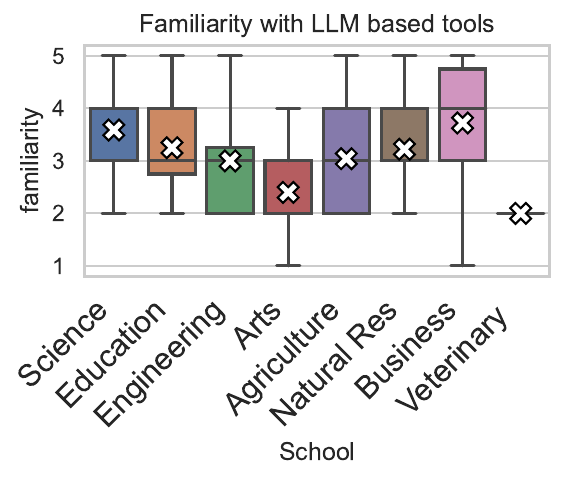}}
    \subfloat[]{
    \label{fig:familiarity_with_llms_by_age}
    \includegraphics[width=0.48\columnwidth]{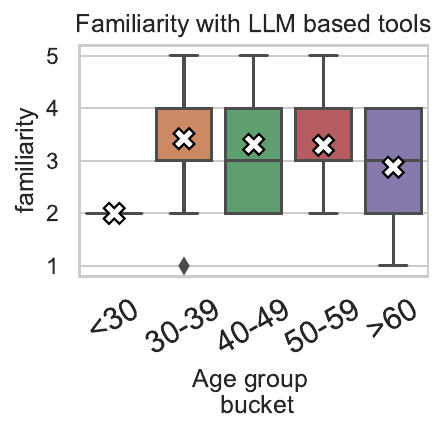}}
    \caption{Familiarity by (a) school and (b) age-group.}
    \label{fig:familarity}
\end{figure}

\subsection{RQ1: How aware are educators of Generative AI-based tools across various departments?}

To answer this question, we asked each survey participant about their familiarity and usage habits of these tools and followed up in an interview with questions about their usage habits, sources of introduction, etc.

Our survey revealed that most educators have at least heard of these tools or tried them. More than 40\% of the faculty members said they use them at least periodically or regularly. 
While no significant difference was found across various age brackets and tenure lengths, the familiarity varied by school (Figure ~\ref{fig:familarity}). The College of Science and School of Business have the highest familiarity overall, while the College of Arts affiliated educators were the least familiar. 

We followed up in the interview on how or in what context the educators were introduced to these tools. Table \ref{tab:discovery-sources} shows the discovery source of these Generative AI-based tools among the educators. Through the interviews, we discovered multiple instances where faculty members who follow the development of these tools more closely held formal or informal workshops to inform their colleagues of these developments. Among those interviewed, 19\% had a technical understanding of Generative AI and LLMs, while others had only a basic understanding. 38\% of the interviewees were very aware that they lacked technical understanding of the tool.

\begin{table}[h]
\centering
\begin{tabular}{|l|c|}
\hline
Discovery Source & Proportion \\
\hline
News & 33.33 \% \\
Peers & 16.67 \% \\
Work/Training & 16.67 \% \\
Family Member & 11.00 \% \\
Social Media & 11.00\%  \\
Others or unsure & 11.33 \% \\
\hline
\end{tabular}
\caption{Discovery sources of Generative AI tools.}
\label{tab:discovery-sources}
\end{table}

\subsection{RQ2: What are educators' perceptions and sentiments about these AI tools?}

Next, we explored the educators' attitudes and sentiments towards these AI tools. We used the answers to the following questions for this:
\begin{enumerate}
    \item AI tools like ChatGPT and Bard should be allowed and integrated into education. (beIntegrated)
    \item I think the AI tools like ChatGPT and Bard should be banned in all academic settings. (beBanned)
\end{enumerate}

We used the following equation to calculate the sentiment and ensure the value is between 1 and 5:
$$ Sentiment = \frac{{\text{{beIntegrated}} + (6 - \text{{beBanned}})}}{2}  $$


\begin{figure}
    \centering
    \subfloat[]{
    \label{fig:sentiment-by-school}
    \includegraphics[width=0.48\columnwidth]{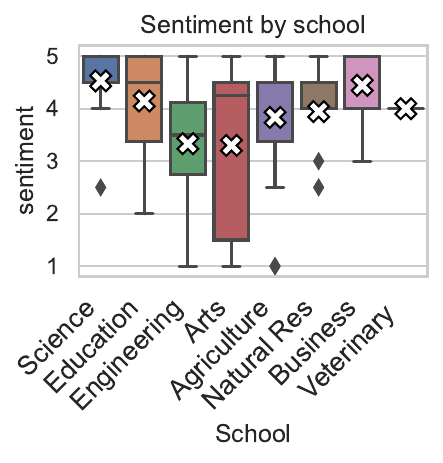}}
    \subfloat[]{
    \label{fig:sentiment-by-age-group}
    \includegraphics[width=0.48\columnwidth]{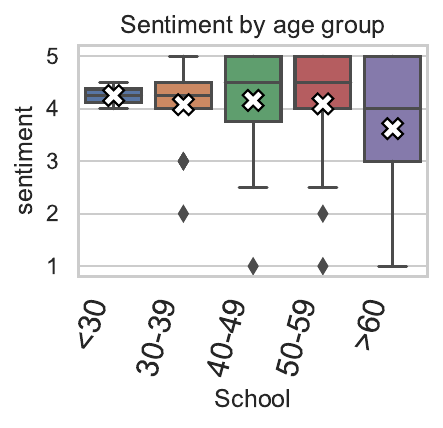}}
    \caption{Sentiment by (a) school and (b) age-group.}
    \label{fig:sentiment}
\end{figure}

The overall sentiment towards these tools is positive, with a mean of 3.99. The median sentiment is 4.5, and the third quartile is 5. Only 12\% of educators had worse than average sentiment ($\text{{sentiment}} < 3$).
Figure \ref{fig:sentiment-by-school} shows the distribution of sentiment by school. Following the familiarity trend, the College of Science and School of Business have the most positive sentiment, while the College of Arts has the lowest.

Figure \ref{fig:sentiment-by-age-group} shows the distribution of sentiment by age group. While the overall mean sentiment is not different across age groups, the inter-quartile range becomes larger for the older age group.


We also asked about their initial impression as well as the change in impression since the first encounter. Most of the respondents, especially from outside the computer science department, used words like "amazed" or "mind-blown" to describe the initial impression.  We saw more than 56\% of interviewees grow more positive, 38\% stayed the same and only 6\% grew more negative.

\subsection {RQ3 : What factors contribute to variations in teachers' attitudes toward AI language model? }

\subsubsection{Pedagogical Practices}

In the survey, we asked the instructors questions about their teaching methodologies like lectures, labs and hands on experiments, discussions etc. as well as the testing methodologies they employ. There was no significant difference discovered between the perception about these AI tools in relation to their pedagogical practices. Comparing both the kinds of questions teachers use for their assignments and test as well as their teaching style, there were no statistically significant differences. We also followed up in the interview about how they see the need to adapt their pedagogical practices to address these new developments.

One of the most-repeated themes was that educators were more receptive of using a tool in advanced classes where students have already acquired the fundamentals of their discipline. 

\begin{displayquote}
          \textit{I have no problem with students using it in my advanced class. In fact, I don't mind that at all, might even encourage it. However, if they use it in the [Intro CS class], they are not going to learn anything.}

\end{displayquote}

This quote from a computer science professor is one of many who indicated they are more positive toward adapting these technologies in higher level classes.

\subsubsection{Identifying contributing features}

Next, we delved into the process of identifying the key contributing features that influence educators' attitudes toward Generative AI and LLMs. To accomplish this, we employed regression analysis and utilized the LASSO (Least Absolute Shrinkage and Selection Operator) technique for feature selection. Through this analysis, we aimed to uncover the most significant factors that play a role in shaping educators' attitudes in order of importance.

The analysis yielded a list of features along with their corresponding coefficients, shedding light on the relative impact of each feature. Table \ref{tab:factors} shows the most important factors that influence teacher's sentiment about Generative AI, listed in ranked order.

\begin{table}[h]
\begin{tabular}{c c c}
\hline
\textbf{Rank} & \textbf{Factor} & \textbf{Effect} \\
\hline
1 & Benefit outweigh risks & Positive \\
2 & Enhances the quality of education & Positive \\
3 & Can easily be integrated & Positive \\
4 & Decreases critical thinking & Negative \\
5 & Increase cheating and dishonesty & Negative\\
 \hline
\end{tabular}
\caption{Top 5 Factors affecting sentiments ranked}
\label{tab:factors}
\end{table}

 It is evident that factors related to risk-to-rewards ratio, quality enhancement, and ease of integrating AI tools are among the most influential in shaping positive attitudes. Conversely, concerns about loss of creativity and potential for cheating and dishonesty impact attitudes negatively.

Moreover, we extended our analysis to different regression techniques, including Linear Regression, Random Forest, Gradient Boost, and XGBoost. The mean squared errors (MSE) obtained ranged between 0.4 and 0.5, indicating a reasonable level of predictive accuracy using these features. 


Overall, our analysis unveils a hierarchy of factors that significantly contribute to educators' attitudes toward Generative AI and LLMs. These insights can guide educational practitioners, policymakers, and researchers in understanding the intricate interplay of factors that shape attitudes, thereby facilitating informed decision-making and effective implementation strategies.




\subsection {RQ4:  How do the attitudes and perceptions of CS educators differ from those from different departments?}

\begin{figure}[th]
\includegraphics[width=8cm]{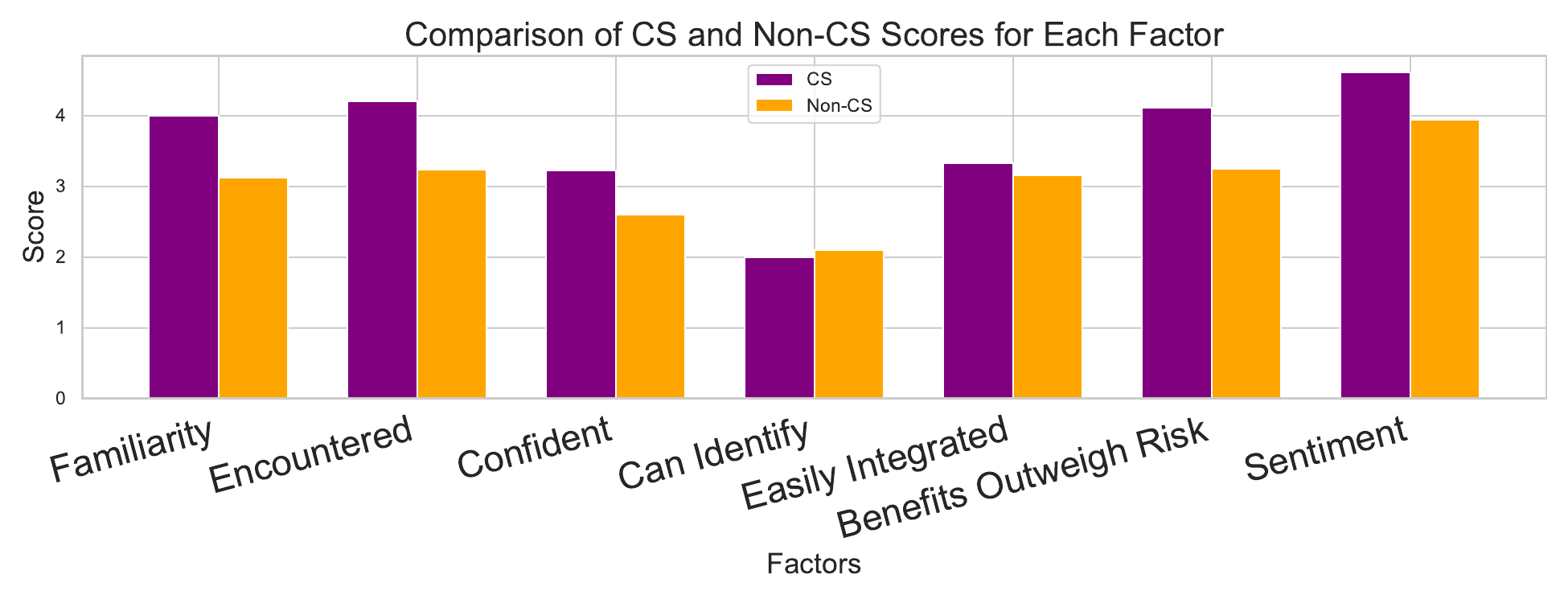}
\caption{Mean response between CS and non-CS instructors.}
\label{fig:cs-noncs}
\vspace{-1em}
\end{figure}

The survey included 9 Computer Science participants out of 116 total, and 6 out of 36 interviewees were from the Computer Science department. In terms of understanding, 83\% of Computer Science instructors had technical understanding compared to only 10\% of Non-CS. In terms of familiarity, as shown in Figure~\ref{fig:cs-noncs}, the CS faculty members reported higher mean familiarity$(M=4, SD=.71)$ as compared to non-CS$(M=3.15, SD=1.08); t(113)=3.28, p = .007$. The majority of CS respondents were confident that their students have used the tools $(M=4.22, SD=1.09)$ while most non-CS $(M=3.23, SD-1.19)$ respondents were unsure, $t(113)=2.57, p = .028$

While we see that the Computer Science instructors had more technical understanding, they have a very similar level of confidence as to whether these new tools can be integrated into education. Similarly, the Computer Science faculty members were even less confident than other faculty members in identifying content generated by AI. This could be because of the nature of assignments (coding in CS vs. more creative writing), or simply because non-CS instructors are over-estimating their confidence. Additionally, some non-CS instructors who haven't used these tools were very surprised when shown an AI generated answer to their questions at the end of the interview.  In terms of overall sentiment, computer science instructors had higher mean sentiment $(M=4.61, SD=0.41)$ as compared to other instructors $(M=3.95, SD=1.11)$ with $t(113)=3.71, p < .001$.

During the interviews, it was observed that CS (Computer Science) instructors were less caught off guard and not as mesmerized by the capabilities of these tools as many Non-CS instructors were. This may be attributed to their gradual exposure to such technologies. Many CS instructors mentioned tools such as GPT, GPT-2, Codex, and Github Copilot, but the most common exposure for Non-CS instructors was to the ChatGPT (GPT-3.5 Turbo) model. CS instructors expressed their view of these tools as a change in approach rather than the end of a topic. One CS professor said:
\begin{displayquote}
  \textit{ "Think how many jobs [no-code solutions like] SquareSpace or Wix.com killed, but our web development class is thriving. I am not worried about it,"}  
\end{displayquote}
On the other hand, some non-CS instructors expressed concern about their work or expertise being valued less.

\subsection {RQ5: What are the biggest opportunities and concerns identified by the educators? }

\begin{table*}[h!]
\begin{tabular}{c c c c}
\hline
\textbf{Opportunities} & \textbf{Percentage} & \textbf{Concerns} & \textbf{Percentage} \\ \hline
Boosts efficiency & 73\% & Potential for cheating & 38\%  \\
Thought Starter / Ideas generator / Springboard & 68\% & Potential to stifle creativity & 38\% \\
Information in fingertips & 53\%  &   Concern about focus on product over process & 36\%  \\
Automate mundane tasks & 53\% & Incorrect or fabricated results &  27\% \\
Personalized teaching/24 hours TA access & 31\% &  Equity and access & 38\%  \\
 \hline
\end{tabular}
\caption{Biggest opportunities and concerns identified by instructors}
\label{tab:positives-negatives}
\end{table*}

In the followup interview, we asked the educators to discuss the biggest opportunities and challenges they see regarding adaptations of these tools in education. Table \ref{tab:positives-negatives} shows the biggest opportunities and challenges identified by educators regarding these generative AI based tools. 

Where there were a number of positives and negatives discussed, even the most frequent concern is discussed only 38\% of time while there are four opportunities discussed over half the time, supporting the idea that educator attitudes are generally more oriented toward opportunities than concerns.

\subsection{Additional observations}
The survey results revealed a notable level of enthusiasm and optimism among faculty members concerning the integration of generative AI tools and Language Learning Models (LLMs) in education. A Business instructor said, \textit{"This is just like the internet in 90s. This is going to change everything."} One of the significant advantages recognized by respondents is the potential for automating mundane and repetitive tasks, freeing up valuable time for educators to focus on more meaningful aspects of teaching. Some educators reported their own creative use of AI for help in grading assignments, generating personalized feedback, creating test questions and even finding flaws and biases in students' arguments.  The notion of using AI as a personal tutor, catering to individualized learning needs, also garnered enthusiasm among respondents. One Arts instructor stated:

\begin{displayquote}
         \textit{I require them to use AI to complete their assignment and submit their prompts, as well as all outputs.}
\end{displayquote}

AI tools are perceived as valuable aids in the creative process. Faculty members acknowledged the utility of AI in generating innovative ideas and offering fresh perspectives on complex concepts. By serving as a tool to bounce ideas off of, AI can challenge conventional approaches, encouraging educators to explore novel teaching methods and content delivery strategies. An instructor in Education said:
\begin{displayquote}
 \textit{[Generative AI] has been a lifeline for for people with learning disorders, or those who need little extra help.}
\end{displayquote}
However, amidst the excitement, several unresolved questions and concerns were highlighted by the faculty members. One major concern pertains to the effective testing of students when AI tools are employed. Traditional testing methods may not adequately assess students' critical thinking and problem-solving skills when assisted by AI, as expressed by an Engineering instructor:

\begin{displayquote}
 \textit{I don't know what to test on anymore. I have no idea how to distinguish a genuine assignment with AI generated."}
\end{displayquote}

Additionally, detecting plagiarism and ensuring academic integrity in an AI-driven learning environment poses a challenge. Moreover, the potential loss of creativity in a heavily AI-driven learning environment sparked debates among faculty members. Another prominent issue is the challenge of combating misinformation or fabricated information. Balancing the use of AI tools while preserving and nurturing students' creativity and originality is an ongoing concern.



\subsection{Limitations}
This survey was self reported, so it has the inherent self-reporting bias. Additionally, the survey was done in one university across different departments, so there could be variation in different institutions. This is also a fast-moving subject, and all the data reflects responses during May and June of 2023.

\section{Conclusions}
While some recent work has cast doubt on whether AI-based tools will or even should become integrated within classrooms \cite{lau2023ban, ko2023more}, our findings reveal that educators are already seeing more positives than negatives.
There is a general consensus from the survey and interview that these generative AI-based tools are going to be part of our education system, and being able to quickly adapt to this new reality sets the direction. While it may not be surprising that educators are aware of AI tools and are becoming more positive, this study's contribution primarily lies in identifying the factors that affect such an environment. Such information can help develop the right policies, conduct necessary training, and provide necessary resources so that we can take advantage of these tools while minimizing risks.

While the potential benefits are promising, it is crucial to navigate the complexities carefully and thoughtfully to ensure an inclusive, equitable, and effective learning experience for all students in the AI era. 
A larger study, encompassing a bigger sample-size, can help generalize these finding. This study also shed light on numerous big-picture philosophical questions that merit further exploration. Fundamental questions about the nature of teaching and learning in the context of AI tools need to be addressed. Existing uncertainty surrounding AI tools and LLM-based technologies in education calls for open dialogues and collaborations between researchers, educators, education policymakers, and technology developers. Together, they can address the emerging challenges, assess ethical considerations, and collectively shape the responsible integration of AI in education.

\bibliographystyle{ACM-Reference-Format}
\bibliography{sample-base}

\end{document}